\documentclass[11pt,a4paper]{article}
\usepackage[hyperref]{acl2021}
\usepackage{times}
\usepackage{latexsym}
\usepackage{linguex}
\usepackage{balance} 
\usepackage{flushend}

\usepackage{microtype}
\usepackage{enumitem}
\usepackage{graphicx}
\usepackage{flushend}

\let\oldbibitem\bibitem
\def\bibitem{\vfill\oldbibitem}

\aclfinalcopy 

\title{An Analysis of the Recent Visibility\\ of the SigDial Conference}

\author{  Casey Kennington \\
  Computer Science\\ Boise State University\\
  \texttt{caseykennington@}\\ \texttt{boisestate.edu} \\ \And
  McKenzie Steenson \\
  Computer Science\\ Boise State University\\
  \texttt{mckenziesteenson@}\\ \texttt{u.boisestate.edu} }

\date{}

\begin{document}
\maketitle
\begin{abstract}
Automated speech and text interfaces are continuing to improve, resulting in increased research in the area of dialogue systems. Moreover, conferences and workshops from various fields are focusing more on language through speech and text mediums as candidates for interaction with applications such as search interfaces and robots. In this paper, we explore how visible the SigDial conference is to outside conferences by analysing papers from top Natural Langauge Processing conferences since 2015 to determine the popularity of certain SigDial-related topics, as well as analysing what SigDial papers are being cited by others outside of SigDial. We find that despite a dramatic increase in dialogue-related research, SigDial visibility has not increased. We conclude by offering some suggestions.
\end{abstract}

\section{Introduction}
The past five years have been transformative for many computational research areas, in many ways due to access to large datasets, compute power, and easy-to-use programming libraries that leverage neural network architectures. The Association for Computational Linguistics (ACL) conference, the flagship conference for Natural Language Processing (NLP) research, and related annual conferences have seen dramatic increases in paper submissions. For example, 2020 had 3,429 paper submissions, 2019 had 2905. Lowering barriers to access the community for researchers and practitioners is a welcome direction for the field. 

Many sub-topics within NLP have seen high representation in research activity over the years (e.g., machine translation) whereas otters have not, and one sub-area that traditionally has not had high representation of paper submissions is dialogue. However, in 2020 \emph{Dialogue and Interactive Systems} was the topic that received the second-highest number of submissions, and for good reason.\footnote{\url{https://acl2020.org/blog/general-conference-statistics/}} An agent that can understand, represent, process, and produce natural language, and then perform a wide range of tasks related to the language it can understand, and that can converse about a diverse number of topics would represent an impressive amount of scientific progress since the beginning days of computation. The practical implications also have potential impact: being able to converse with any machine about its capabilities, errors, or even about the weather would finally give machines an interface that is close to the most natural for humans: spoken language in an interactive setting. 

This area of research has its own set of open research challenges. For example, multi-turn discourse, turn-taking, conversational grounding, handling disfluencies, incremental processing, language understanding, dialogue managment, and language generation to name a few. Furthermore, if speech, not just text, is the primary medium of communication, then speech recognition and speech synthesis must be included as active areas of research. \emph{The Special Interest Group on Discourse and Dialogue} (SigDial), which has held annual meetings (first workshops, then conferences) since 1998 is the official special interest group for these kinds of topics of discourse and dialogue for ACL and the International Speech Communication Association (ISCA). 

The goal of this paper is to explore the growth of discourse and dialogue research and the visibility of SigDial in that growth.

\section{Data Collection}

In order to track the impact of SigDial on the broader community, We chose the top three rated NLP conferences ACL from years 2015-2020 \cite{acl-2015-association,acl-2016-association,acl-2017-association,acl-2018-association,acl-2019-association,acl-2020-association}; EMNLP (Emprirical Methods of NLP) from 2015-2020 \cite{emnlp2015,emnlp2016,emnlp2017,emnlp2018,emnlp2019,emnlp2020}; and NAACL (North American ACL) from years it was held during the same time frame, namely 2015, 2016, 2018, and 2019 \cite{naacl-2015-2015,naacl-2016-2016,naacl-2018-2018,naacl-2019-2019}.\footnote{{These are top rated for NLP according to csrankings.org}}  and focused on accepted papers from 2015 to 2020 to each of these conferences. We also collected long papers from the Conversational User Inverface (CUI) conference from 2019 and 2020 (the conference began in 2019) as well as two recent related ACL workshops: \emph{Workshop on Search-Oriented Conversational AI} which we denote as SearchAI \cite{ws-2018-2018-emnlp} and \emph{NLP Workshop on Conversational AI} which we denote as ConvAI \cite{ws-2019-nlp-conversational}. For a baseline comparison, we also processed the SigDial proceedings for the years 2015-2020 \cite{sigdial-2015,sigdial-2016,sigdial-2017,sigdial-2018,sigdial-2019,sigdial-2020} 


For each accepted paper, we performed the following: convert the PDF to text, then search for the following keywords (and phrases):

\begin{itemize}[noitemsep]
    \item \emph{discourse}; we denote this as \emph{disc} in figures and tables below.
    \item \emph{interaction} (this could include other areas of research that are related to, but distinct from dialogue research, including non-verbal interaction); we denote this as \emph{int} below.
    \item \emph{chat} (a prefix for chat-based systems)
    \item \emph{dialog} (\emph{dialog} and \emph{dialogue} are both common spellings; the former is a prefix of the latter)
    \item \emph{conversational} (conversational-AI is the term given to systems that interact, including dialogue systems); we denote this as \emph{conv} below.
    \item \emph{sigdial} (to capture references often use the SigDial nickname)
    \item \emph{special interest group for discourse and dialogue} (to capture references made to papers accepted to a SigDial conference); we denote this as \emph{sig-full} below. 
\end{itemize}

We counted occurrences for each above keyword for each PDF. This resulted in a dataset with the paper title, year, author list, as well as the counts for each of the above keywords. We removed PDFs from the dataset that had processing errors. Statistics for all venues (aggregated for all years) are in Table~\ref{tab:pdf_stats}. This resulted in a total of 6,184 papers for 24 venues across 6 years. 

\begin{table}
\centering
{
\footnotesize
\addtolength{\tabcolsep}{-3pt}   
\begin{tabular}{lccc}
\hline
venue & \# papers & \# removed & \% removed\\
\hline
ACL & 2170 & 124 & 5.7\% \\
EMNLP & 2709 & 171 & 6.31\% \\
NAACL & 982 & 13 & 1.32\%  \\
SigDial & 294 & 4 & 1.36\% \\
CUI     & 15 & 0 & 0\% \\
SearchAI & 13 & 0 & 0\% \\
ConvAI & 16 & 0 & 0\% \\
\hline
total & 6184 & 312 & 5.06\% \\
\end{tabular}
\caption{ \label{tab:pdf_stats} Numbers of processed papers (and papers removed) from the final data set used for analysis.}
}
\end{table}

\section{Analyses}

In this section, we describe how we processed the papers and the resulting analyses. 

\paragraph{Analysis of Keywords}

In this section, we show the average per-paper mentions of specific keywords (\emph{discourse, interaction, chat, dialog, conversational, sigdial, the special interest group for discourse and dialogue}) for SigDial, ACL, EMNLP, NAACL, CUI, and two related workshops. We offer figures in the form of bar charts to make the numbers easily comparable across time for each conference (though note the y-axis ranges are different for each); tables with specific corresponding numbers can be found in the Appendix. 

\begin{figure}
  \includegraphics[width=1.0\linewidth]{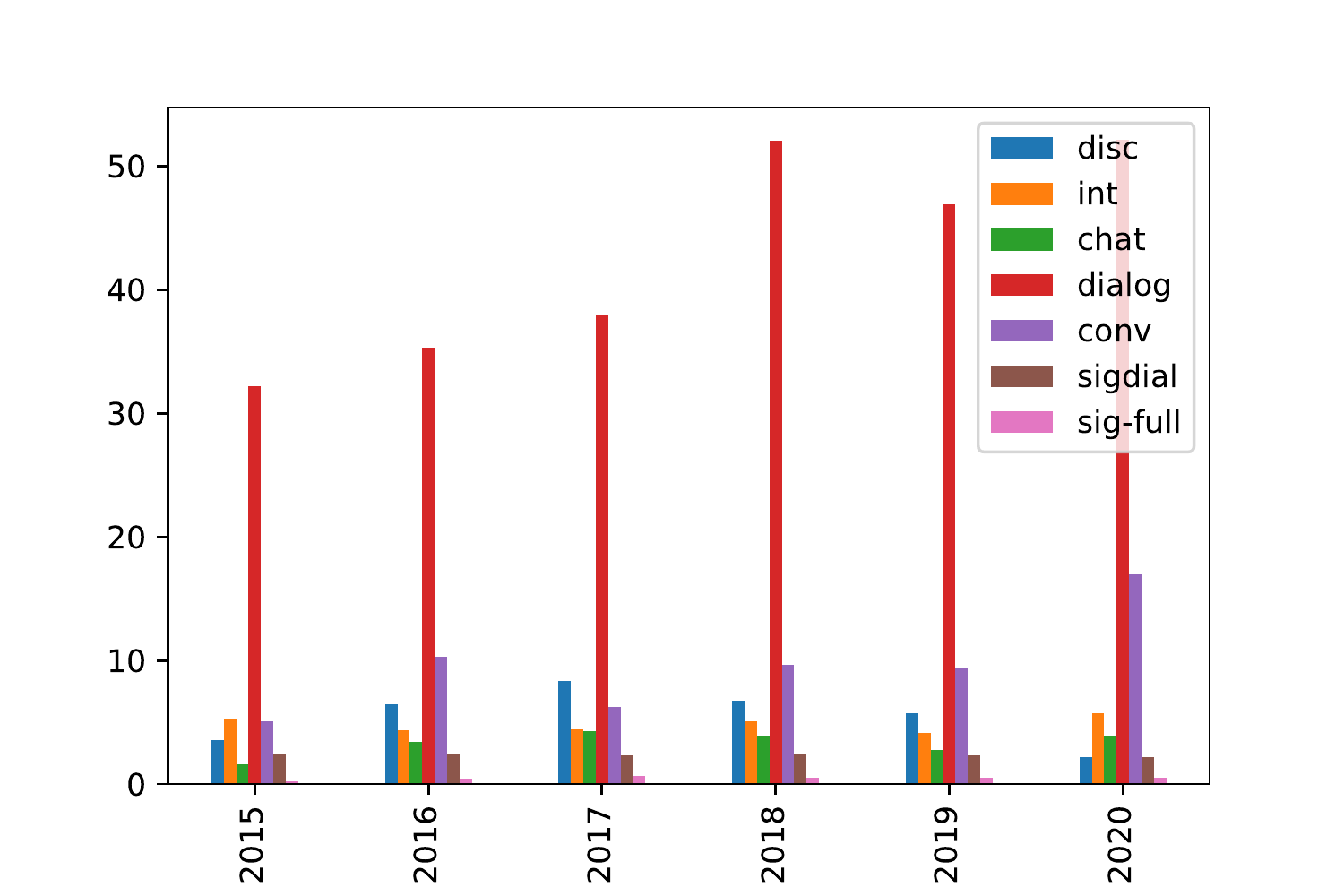}
  \caption{Counts of mentions of keywords in SigDial papers from 2015-2020.}
  \label{fig:sigdial_stats}
\end{figure}

\begin{figure}[h]
  \includegraphics[width=1.0\linewidth]{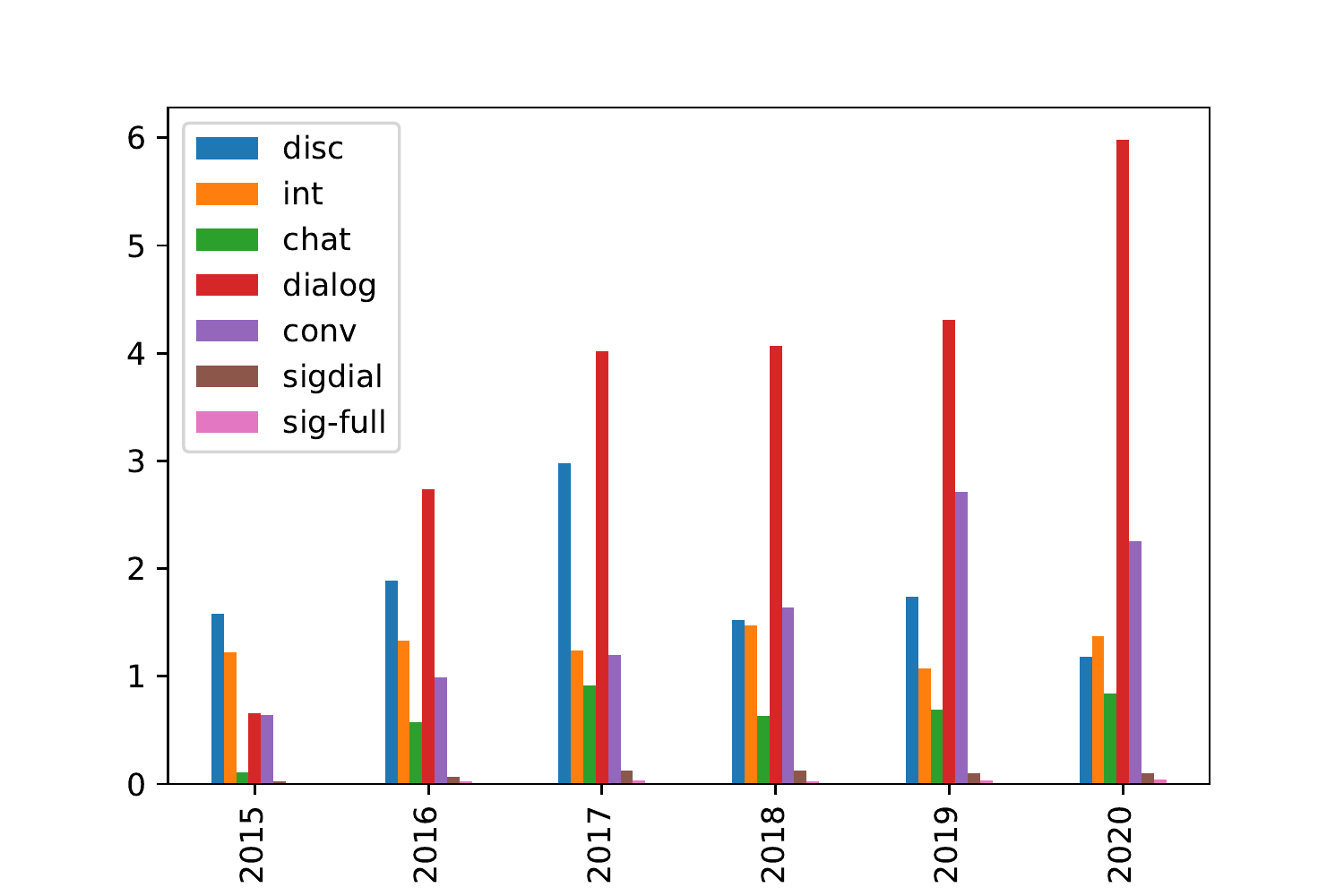}
  \caption{Average counts  of mentions of keywords in ACL papers from 2015-2020.}
  \label{fig:acl_stats}
\end{figure}

\begin{figure}[h]
  \includegraphics[width=1.0\linewidth]{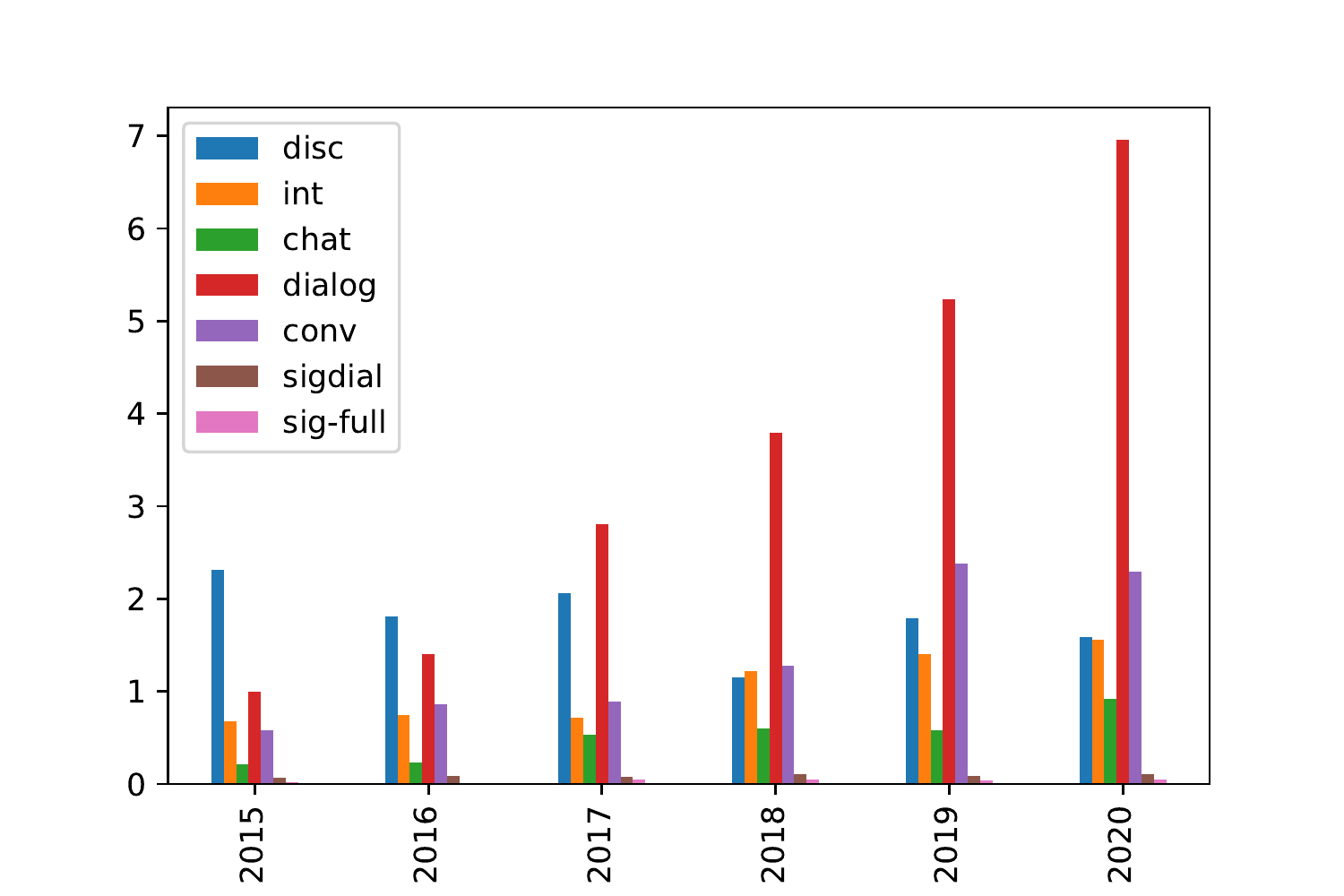}
  \caption{Average counts  of mentions of keywords in EMNLP papers from 2015-2020.}
  \label{fig:emnlp_stats}
\end{figure}

\begin{figure}[h]
  \includegraphics[width=1.0\linewidth]{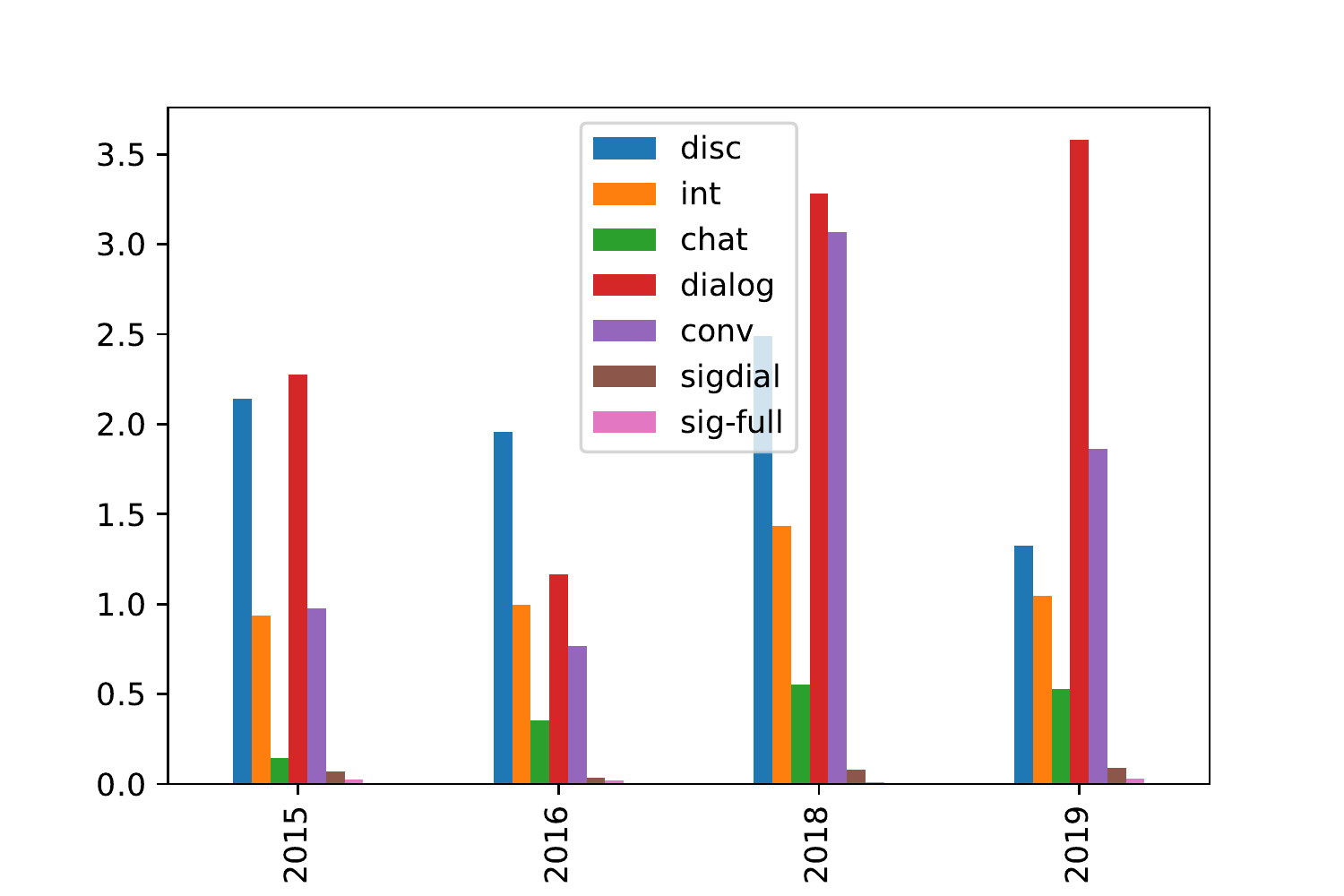}
  \caption{Average counts  of mentions of keywords in NAACL papers from 2015-2020.}
  \label{fig:naacl_stats}
\end{figure}

Figures~\ref{fig:acl_stats}, \ref{fig:emnlp_stats}, \ref{fig:naacl_stats} show the average per-paper mentions of the keywords in question for ACL, EMNLP, and NAACL from 2015-2020. These figures tell a consistent story, albeit a noteably different story from the SigDial numbers.

The first thing of note is the steady increase of the use of the word \emph{dialog[gue]} over time for the *ACL conference. A regression analysis shows that \emph{dialogue} increases by double nearly every year (slope=0.9) for ACL (with similar numbers for EMNLP, less for NAACL). The word \emph{chat} has gone up less drastically, but it too has shown a steady increase over time (slope=0.11), and \emph{conversational} has had a slow steady increase as well (slope=0.4). This is a somewhat surprising result, as it was our impression that "conversational AI" is a commonly used word to denote certain types of dialogue systems, but that doesn't appear to be a word commonly used in the NLP community. We conjecture that this could be due to the accepted definition of ``conversation" connotes the kinds of spoken or text interactions that aren't tied to a task, whereas dialogue systems traditionally have been tied to a specific task. Keywords \emph{discourse} and \emph{interaction} have had an overall negative slope (-0.1 and -0.006, respectively), whereas the \emph{sigdial} and \emph{special interest group for discourse and dialogue} have positive slopes of 0.04 and 0.005, respectively. We don't show figures here, but found similar results for the two workshops (see Tables~\ref{tab:averages} and \ref{tab:counts} for details on averages and counts for all conferences and workshops considered here). 

\begin{figure}[h]
  \includegraphics[width=0.9\linewidth]{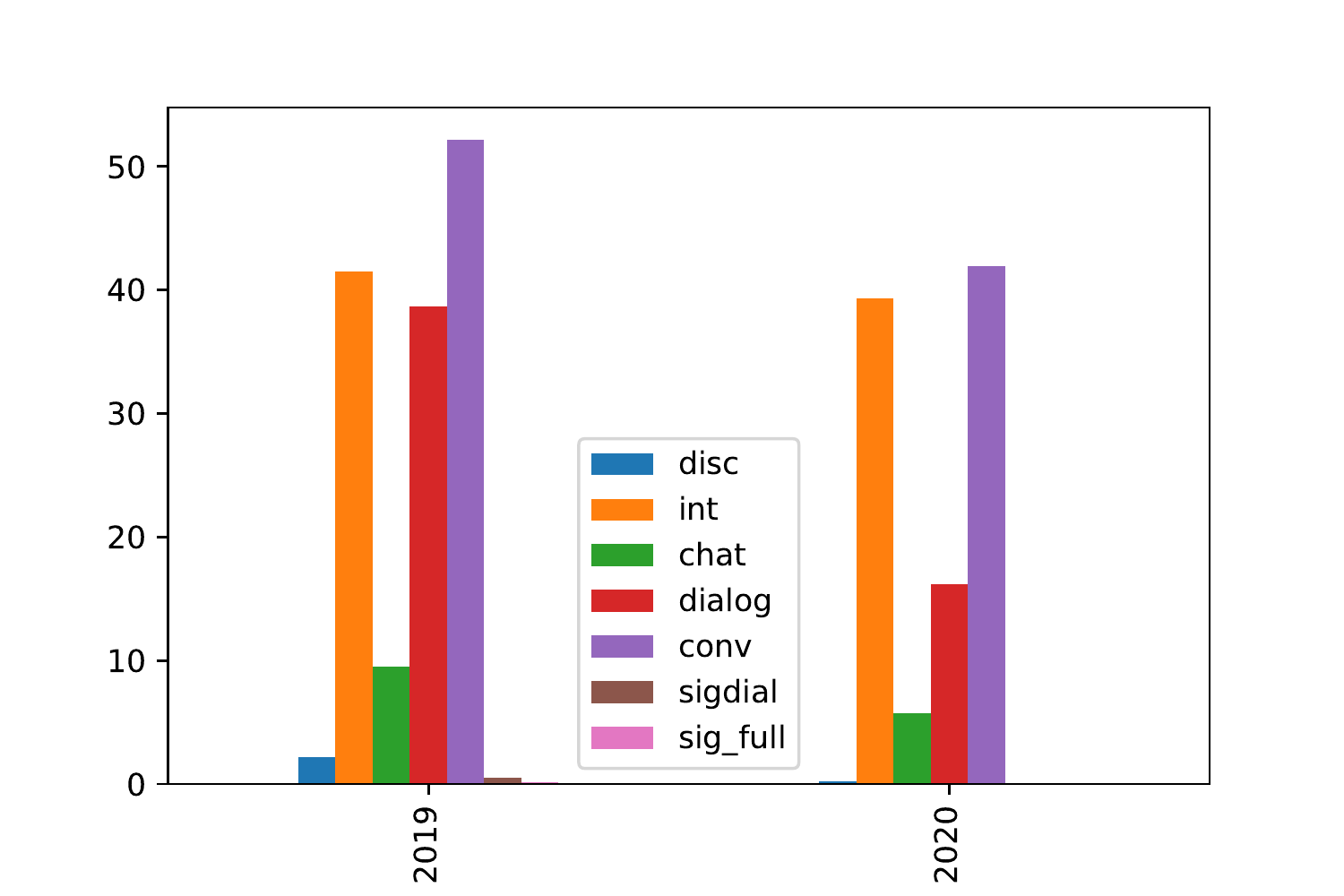}
  \caption{Average counts of mentions of keywords in CUI papers from 2019-2020.}
  \label{fig:cui_stats}
\end{figure}

Figure~\ref{fig:cui_stats} shows the statistics from the CUI conference, 2019-2020. This is an important comparison because the conference is outside of the NLP community and thus represents a new established conference. More focus at CUI appears to be on the interface design and mostly focuses on Conversational AI. The average counts show that Conversational AI is the most commony term, though \emph{interaction} and \emph{dialogue} are common. From 15 long papers over two years, the conference only makes 4 citations to SigDial papers. 

\paragraph{Analysis of References} 

In this section we focus on the \emph{sigdial} and \emph{special interest group on discourse and dialogue} keywords to determie which papers outside conferences have been citing. We processed each PDF, extracting references that cited SigDial using one of the above two keywords. The total number of citations to SigDial for the ACL, EMLNLP, and NAACL conferences together was 735 (from of 6,184 papers); for the two workshops together the count was 24 (from 29 papers). For comparison, the number of refrences to SigDial papers from SigDial conferences of the same timeframe is 821 (from 294 papers; more information on counts can be found in Table~\ref{tab:counts} in the Appendix). 

The most commonly cited paper was \newcite{henderson-etal-2014-second}, which is a report on the Second Dialogue State Tracking Challenge (DTSC), for a total of 46 citations. Two other highly-cited papers were \newcite{williams-etal-2013-dialog} 8 times (the first DTSC), and \newcite{henderson-etal-2014-word} 7 times (a paper describing an approach to state tracking). These three papers account for 8.3\% of all citations to SigDial. The second-highest cited paper was a resource paper (The Ubuntu Dialog Corpus), with 27 citations \cite{lowe-etal-2015-ubuntu}. Two other papers had 7 citations each, one focusing on a discourse-tagged corpus, another focusing on conversational structure of agreement and disagreement \cite{carlson-etal-2001-building,rosenthal-mckeown-2015-couldnt}. One other paper had 6 citations, 5 papers had 5 citations, 7 papers had 11 citations, 11 papers had 3 citations, and the rest had 2 or fewer citations.\footnote{See also \url{https://tinyurl.com/57h934jf} that lists top total found references to some SigDial papers.}

\subsection{Discussion: Challenges in Research for Discourse and Dialogue}

While the general NLP field is shifting to some degree towards dialogue and chat-related research--a welcome trend!--the field neglects what SigDial researchers have uncovered over multiple decades of focus on discourse and dialogue (particularly the latter). This is evidenced by SigDial having fairly low visibility as a conference, despite the rising popularity of the topic. This is a two-edged sword: on one side, it is useful to dispose of old ideas that may no longer be relevant thanks to recent advancements (e.g., deep learning, new datasets), but on the other side, scientific knowledge is re-discovered. Moreover, lack of knowledge of discourse and dialogue phenomena could lead to creation of the wrong kinds of datasets. For example, in 2015, the bAbI dataset was realeased \cite{weston2015towards}, including a dialogue dataset. Subsequent papers that made use of the data to train dialogue models made broad-sweeping claims of understanding language until \newcite{shalyminov-etal-2017-challenging} challenged the notion that the bAbI Dialogues Dataset and corresponding tasks reflected the kinds of real-world conversations that people have. By inserting common conversational artifacts like hesitations and repetitions, which did not exist in the original data,  the state-of-the-art, end-to-end approaches to the bAbI tasks up to that point lacked the ability to generalize to common and natural language phenomena found in spontaneous dialogue data.

While it is true that developing working systems requires ample data and engineering, handling various dialogue phenomena together in a single system is \emph{not} low-hanging fruit for research. It is our observation that dialogue artifacts like repetitions and disfluencies (which are not common in text) are not well-known in the broader research community. Discourse is equally challenging, though from the analyses above, it is clear that discourse is very under-studied, even from within SigDial.

\section{Conclusion: Implications for SigDial}

Dialogue is a hot topic, but the NLP field is potentially spending unnecessary time re-learning the lessons that the SigDial community has learned over the past two decades: that spoken interfaces are difficult to design (and are sometimes not the solution \cite{balentine2007s}) due to the nuances that humans exhibit and expect when using language as a medium of communication. The SigDial community has a big potential here for growth and inclusivity for researchers and practitioners. There clearly is interest in the community; in 2020, like many other conferences, the SigDial conference was fully online, allowing broad participation. Whereas SigDial usually sees 100-150 participants at its annual conference, over 600 participants took part in SigDial in 2020.\footnote{\url{https://t.co/fivuw0HtnN?amp=1}} Further evidence is the creation of related workshops (e.g., the two listed above) and conferences such as Conversational User Interfaces, which is an effort to incorporate spoken and text interaction into computational interfaces.

If research in dialogue is to continue in a forward direction where researchers are not re-learning aspects of dialogue already well-known to the SigDial community, more outreach by community members needs to take place. As evidenced by the high citations to the DTSC efforts, providing more challenges to the larger community could have impact. Also, more outreach to other communities where a language (speech or text) interface would be beneficial, such as web search, recommender systems, or robotics.


\balance
\bibliographystyle{acl_natbib}
\bibliography{refs}

\newpage
\section*{Appendix}

\begin{table}[h]
\centering
{
\footnotesize
\addtolength{\tabcolsep}{-3pt}   
\begin{tabular}{|l|c|c|c|c|c|c|}
\hline
\textbf{ACL} &   \textbf{2015} &   \textbf{2016} &   \textbf{2017} &   \textbf{2018} &   \textbf{2019} &   \textbf{2020} \\
\hline
disc     &  1.58 &  1.89 &  2.97 &  1.52 &  1.74 &  1.18 \\
int      &  1.22 &  1.33 &  1.24 &  1.48 &  1.07 &  1.38 \\
chat     &  0.11 &  0.57 &  0.92 &  0.64 &  0.69 &  0.84 \\
dialog   &  0.66 &  2.74 &  4.02 &  4.07 &  4.31 &  5.98 \\
conv     &  0.64 &  0.99 &  1.19 &  1.64 &  2.71 &  2.25 \\
sigdial  &  0.02 &  0.07 &  0.13 &  0.12 &  0.10 &  0.10 \\
sig-full &  0.01 &  0.03 &  0.03 &  0.02 &  0.03 &  0.04 \\
\hline
\textbf{EMLNP} &   \textbf{2015} &   \textbf{2016} &   \textbf{2017} &   \textbf{2018} &   \textbf{2019} &   \textbf{2020} \\
\hline
disc     &  2.32 &  1.81 &  2.06 &  1.15 &  1.79 &  1.59 \\
int      &  0.68 &  0.75 &  0.72 &  1.22 &  1.41 &  1.55 \\
chat     &  0.22 &  0.23 &  0.54 &  0.60 &  0.58 &  0.92 \\
dialog   &  1.00 &  1.41 &  2.81 &  3.80 &  5.24 &  6.96 \\
conv     &  0.59 &  0.86 &  0.89 &  1.27 &  2.38 &  2.29 \\
sigdial  &  0.07 &  0.08 &  0.08 &  0.11 &  0.09 &  0.11 \\
sig-full &  0.02 &  0.01 &  0.05 &  0.04 &  0.04 &  0.05 \\
\hline
\textbf{SIGDIAL} &   \textbf{2015} &   \textbf{2016} &   \textbf{2017} &   \textbf{2018} &   \textbf{2019} &   \textbf{2020} \\
\hline
disc     &   3.57 &   6.43 &   8.32 &   6.74 &   5.72 &   2.17 \\
int      &   5.28 &   4.39 &   4.45 &   5.11 &   4.10 &   5.70 \\
chat     &   1.57 &   3.43 &   4.26 &   3.89 &   2.76 &   3.90 \\
dialog   &  32.18 &  35.29 &  37.91 &  52.11 &  46.92 &  52.15 \\
conv     &   5.05 &  10.27 &   6.21 &   9.68 &   9.40 &  17.00 \\
sigdial  &   2.36 &   2.45 &   2.32 &   2.38 &   2.32 &   2.17 \\
sig-full &   0.23 &   0.43 &   0.64 &   0.49 &   0.50 &   0.50 \\
\hline
\textbf{NAACL} &   \textbf{2015} &   \textbf{2016} &   \textbf{2017} &   \textbf{2018} &   \textbf{2019} &   \textbf{2020} \\
\hline
disc     &  2.14 &  1.96 & & 2.49 &  1.32 &\\
int      &  0.94 &  0.99 & & 1.44 &  1.05 &\\
chat     &  0.15 &  0.35 & & 0.55 &  0.53 &\\
dialog   &  2.27 &  1.17 & & 3.28 &  3.58 &\\
conv     &  0.98 &  0.77 & & 3.07 &  1.86 &\\
sigdial  &  0.07 &  0.03 & & 0.08 &  0.09 &\\
sig-full &  0.03 &  0.02 & & 0.01 &  0.03 &\\
\hline
\textbf{CUI} &    &    &    &   &   \textbf{2019} &   \textbf{2020}   \\
\hline
disc     &   &  &    &   &  2.2 & 0.2 \\
int      &    &  &    &   & 41.5  & 39.3 \\
chat     &    &  &    &   & 9.5  & 5.7\\
dialog   &  &  &    &   &  38.7 & 16.2 \\
conv     &   &  &    &   & 52.17  & 41.9\\
sigdial  &   &  &    &   & 0.5  & 0.0 \\
sig-full &   & &    &   &  0.17 & 0.0 \\
\hline
\textbf{} &   \textbf{ConvAI} &   \textbf{Search} &    &   &   &   \\
\hline
disc     &   0.62 & 2.54 &    &   &   & \\
int      &   1.25 & 4.85 &    &   &   & \\
chat     &   5.75 & 5.62 &    &   &   & \\
dialog   &  28.19 & 30.77 &    &   &   &  \\
conv     &  20.50 & 11.00 &    &   &   & \\
sigdial  &   1.00 & 0.46 &    &   &   &  \\
sig-full &   0.12 & 0.00 &    &   &   & \\
\hline
\end{tabular}
\caption{Average mentions of keywords per paper for ACL, EMNLP, SigDial, NAACL, CUI, the Conversational AI 2019 Workshop (ConvAI), and the Conversational AI for Search (Search) 2018 Workshop. \label{tab:averages} }
}
\end{table}

\begin{table}[h]
\centering
{
\footnotesize
\addtolength{\tabcolsep}{-3pt}   
\begin{tabular}{|l|c|c|c|c|c|c|}
\hline
\textbf{ACL} &   \textbf{2015} &   \textbf{2016} &   \textbf{2017} &   \textbf{2018} &   \textbf{2019} &   \textbf{2020} \\
\hline
disc     &   275 &   437 &   580 &   369 &  1072 &   843 \\
int      &   213 &   308 &   242 &   357 &   660 &   979 \\
chat     &    19 &   132 &   179 &   154 &   427 &   598 \\
dialog   &   114 &   633 &   784 &   984 &  2654 &  4259 \\
conv     &   112 &   229 &   233 &   396 &  1670 &  1604 \\
sigdial  &     4 &    16 &    25 &    30 &    62 &    73 \\
sig\_full &     1 &     6 &     6 &     6 &    21 &    28 \\
\hline
\textbf{EMLNP} &   \textbf{2015} &   \textbf{2016} &   \textbf{2017} &   \textbf{2018} &   \textbf{2019} &   \textbf{2020} \\
\hline
disc     &   720 &   479 &   666 &   590 &  1127 &  1059 \\
int      &   212 &   197 &   232 &   627 &   886 &  1037 \\
chat     &    67 &    61 &   173 &   306 &   364 &   612 \\
dialog   &   310 &   371 &   907 &  1951 &  3301 &  4642 \\
conv     &   182 &   227 &   288 &   655 &  1497 &  1528 \\
sigdial  &    22 &    22 &    25 &    56 &    57 &    72 \\
sig\_full &     6 &     3 &    16 &    23 &    27 &    33 \\
\hline
\textbf{SIGDIAL} &   \textbf{2015} &   \textbf{2016} &   \textbf{2017} &   \textbf{2018} &   \textbf{2019} &   \textbf{2020} \\
\hline
disc     &   218 &   315 &   391 &   317 &   286 &    87 \\
int      &   322 &   215 &   209 &   240 &   205 &   228 \\
chat     &    96 &   168 &   200 &   183 &   138 &   156 \\
dialog   &  1963 &  1729 &  1782 &  2449 &  2346 &  2086 \\
conv     &   308 &   503 &   292 &   455 &   470 &   680 \\
sigdial  &   144 &   120 &   109 &   112 &   116 &    87 \\
sig\_full &    14 &    21 &    30 &    23 &    25 &    20 \\
\hline
\textbf{NAACL} &   \textbf{2015} &   \textbf{2016} &   \textbf{2017} &   \textbf{2018} &   \textbf{2019} &   \textbf{2020} \\
\hline
disc     &   398 &   354 &  & 508 &   544 &\\
int      &   174 &   180 &  & 293 &   430 &\\
chat     &    27 &    64 &  & 113 &   217 &\\
dialog   &   423 &   211 &  & 670 &  1472 &\\
conv     &   182 &   139 &  & 626 &   765 &\\
sigdial  &    13 &     6 &  &  16 &    36 &\\
sig\_full &     5 &     4 & &    2 &    13 & \\
\hline
\textbf{CUI} &   &  &    &   &   \textbf{2019} &   \textbf{2020}    \\
\hline
disc     &     &   &    &  &13 & 2  \\
int      &     &   &    &  &249 & 393  \\
chat     &     &   &    &  &57 & 57  \\
dialog   &    &   &    & &232  & 162  \\
conv     &    &   &    &  &313 & 419  \\
sigdial  &     &   &    & &3  &  0\\
sig-full &     &   &    & &1  & 0  \\
\hline
\textbf{} &   \textbf{ConvAI} &   \textbf{Search} &    &   &   &   \\
\hline
disc     &    10 & 33  &    &   &   &\\
int      &    20 & 63  &    &   &   &\\
chat     &    92 & 73  &    &   &   &\\
dialog   &   451 & 400  &    &   &   &\\
conv     &   328 & 143  &    &   &   &\\
sigdial  &    16 & 6  &    &   &   &\\
sig-full &     2 & 0  &    &   &   &\\
\hline
\end{tabular}
\caption{Counts of mentions of keywords for ACL, EMNLP, SigDial, NAACL, CUI, the Conversational AI 2019 Workshop (ConvAI), and the Conversational AI for Search (Search) 2018 Workshop. \label{tab:counts} }
}
\end{table}

\end{document}